\documentclass{article} %
\usepackage{iclr2022_conference,times}

\usepackage{hyperref}
\usepackage{url}
\usepackage{mathtools,amssymb}
\usepackage{pgfplots,pgfplotstable}
\pgfplotsset{compat=1.14}
\usepackage{array,colortbl}
\usepackage{xcolor}
\usepackage{algorithm,algorithmicx,algpseudocode}
\usepackage[capitalise]{cleveref}
\usepackage{caption}
\usepackage{graphbox}
\usepackage{placeins}
\usepackage{wrapfig}
\usepackage{subcaption}
\usepackage{etoolbox}
\usepackage{booktabs}       %
\usepackage{microtype}      %
\usepackage{adjustbox}
\usepackage[bottom]{footmisc}

\newcommand{\grad}{\nabla}
\newcommand{\E}{\mathbb{E}}

\newcommand{\Eb}[2]{\E_{#1}\!\left[#2\right]}

\newcommand{\bI}{\mathbf{I}}

\newcommand{\bzero}{\mathbf{0}}

\newcommand{\bc}{\mathbf{c}}

\newcommand{\bx}{\mathbf{x}}

\newcommand{\bz}{\mathbf{z}}

\newcommand{\bepsilon}{{\boldsymbol{\epsilon}}}
\newcommand{\bmu}{{\boldsymbol{\mu}}}

\title{Classifier-Free Diffusion Guidance}

\author{Jonathan Ho \& Tim Salimans \\
Google Research, Brain team\\
\texttt{\{jonathanho,salimans\}@google.com}
}

\iclrfinalcopy %
\begin{document}

\maketitle

\begin{abstract}
Classifier guidance is a recently introduced method to trade off mode coverage and sample fidelity in conditional diffusion models post training, in the same spirit as low temperature sampling or truncation in other types of generative models. Classifier guidance combines the score estimate of a diffusion model with the gradient of an image classifier and thereby requires training an image classifier separate from the diffusion model. It also raises the question of whether guidance can be performed without a classifier. We show that guidance can be indeed performed by a pure generative model without such a classifier: in what we call classifier-free guidance, we jointly train a conditional and an unconditional diffusion model, and we combine the resulting conditional and unconditional score estimates to attain a trade-off between sample quality and diversity similar to that obtained using classifier guidance.\let\thefootnote\relax\footnote{A short version of this paper appeared in the NeurIPS 2021 Workshop on Deep Generative Models and Downstream Applications: \url{https://openreview.net/pdf?id=qw8AKxfYbI}}
\end{abstract}

\section{Introduction}
\label{sec:introduction}

\setlength{\tabcolsep}{0.5pt} %
\begin{figure}[b]\centering
\begin{tabular}{lccr}
\adjincludegraphics[width=0.86\textwidth,Clip={0.5\width} {0.427\height} {0.212\width} {0\height}]{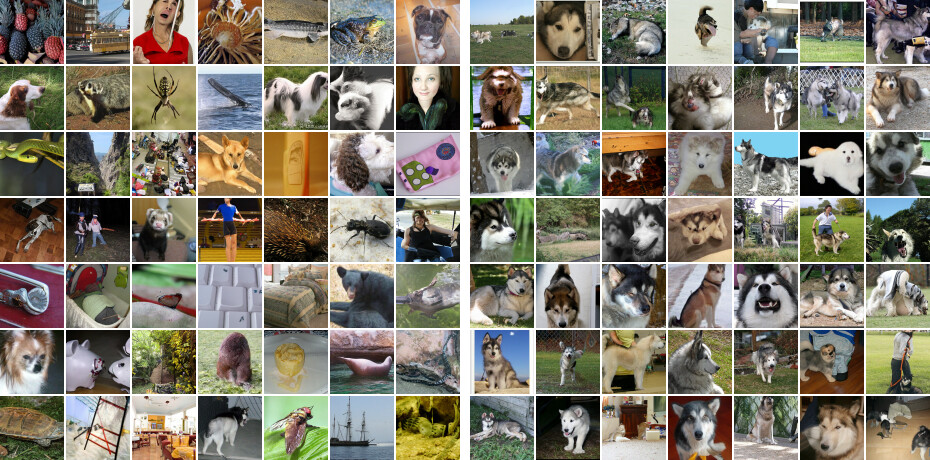} &
\adjincludegraphics[width=0.86\textwidth,Clip={0.5\width} {0.427\height} {0.212\width} {0\height}]{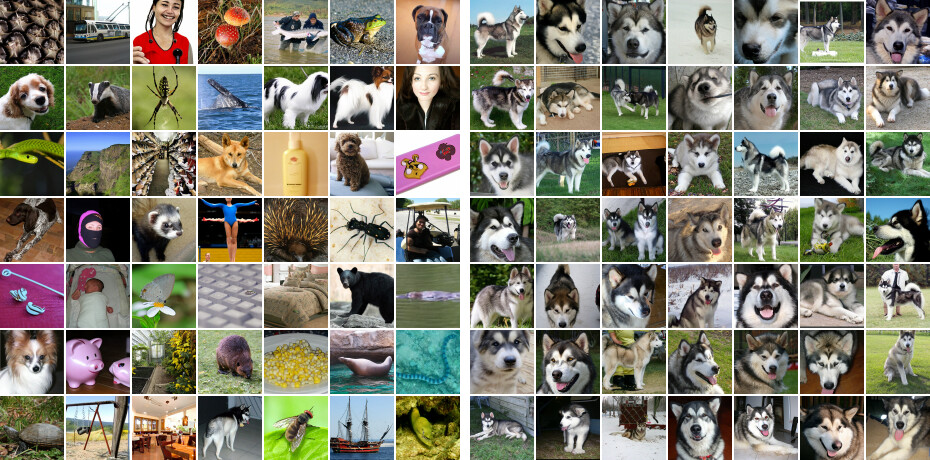} &
\adjincludegraphics[width=0.86\textwidth,Clip={0.5\width} {0.427\height} {0.212\width} {0\height}]{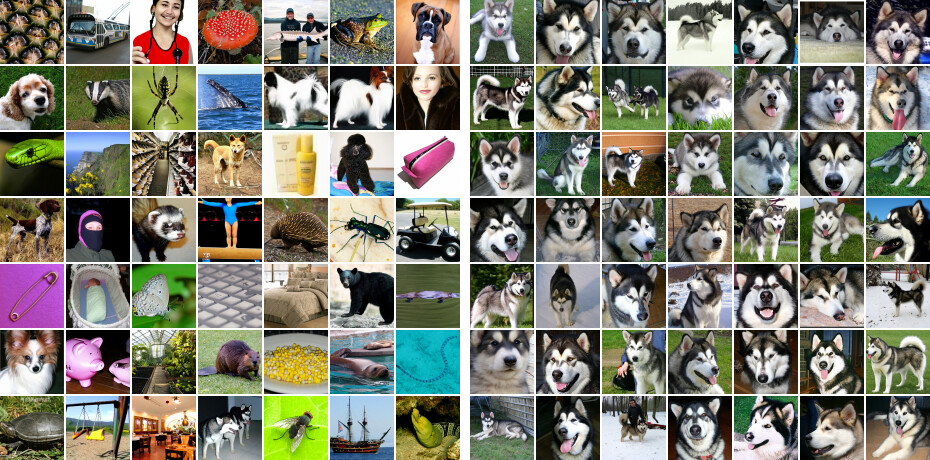} &
\adjincludegraphics[width=0.86\textwidth,Clip={0.5\width} {0.427\height} {0.212\width} {0\height}]{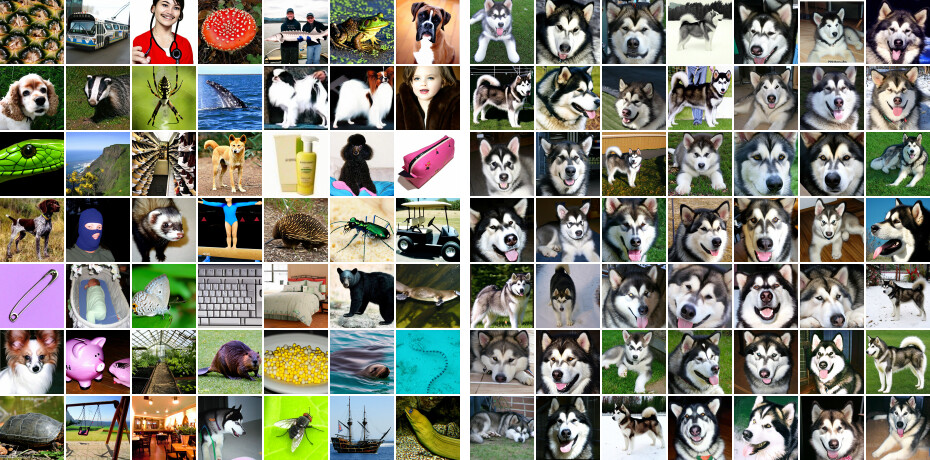}
\end{tabular}
\caption{Classifier-free guidance on the malamute class for a 64x64 ImageNet diffusion model. Left to right: increasing amounts of classifier-free guidance, starting from non-guided samples on the left.}
\label{fig:dog_guidance}
\end{figure}
\setlength{\tabcolsep}{6pt} %

\begin{figure}[t] \centering
\includegraphics[width=\linewidth]{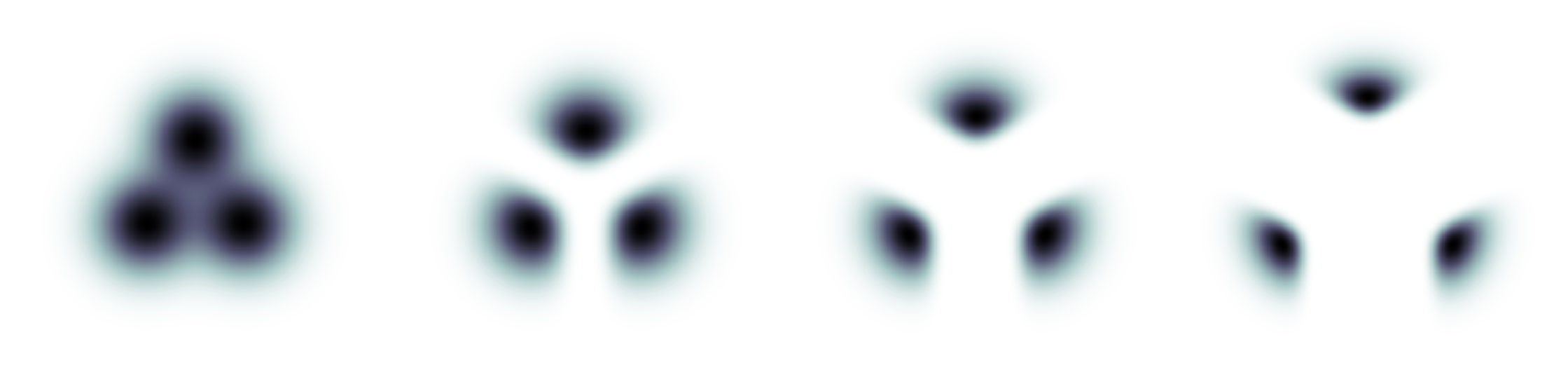}
\caption{The effect of guidance on a mixture of three Gaussians, each mixture component representing data conditioned on a class. The leftmost plot is the non-guided marginal density. Left to right are densities of mixtures of normalized guided conditionals with increasing guidance strength.}
\label{fig:gaussian_guidance}
\end{figure}

Diffusion models have recently emerged as an expressive and flexible family of generative models, delivering competitive sample quality and likelihood scores on image and audio synthesis tasks~\citep{sohl2015deep,song2019generative,ho2020denoising,song2020score,kingma2021variational,song2021maximum}. These models have delivered audio synthesis performance rivaling the quality of autoregressive models with substantially fewer inference steps~\citep{chen2020wavegrad,kong2020diffwave}, and they have delivered ImageNet generation results outperforming BigGAN-deep~\citep{brock2018large} and VQ-VAE-2~\citep{razavi2019generating} in terms of FID score and classification accuracy score~\citep{ho2021cascaded,dhariwal2021diffusion}.

\citet{dhariwal2021diffusion} proposed \emph{classifier guidance}, a technique to boost the sample quality of a diffusion model using an extra trained classifier. Prior to classifier guidance, it was not known how to generate ``low temperature'' samples from a diffusion model similar to those produced by truncated BigGAN~\citep{brock2018large} or low temperature Glow~\citep{kingma2018glow}: naive attempts, such as scaling the model score vectors or decreasing the amount of Gaussian noise added during diffusion sampling, are ineffective~\citep{dhariwal2021diffusion}. Classifier guidance instead mixes a diffusion model's score estimate with the input gradient of the log probability of a classifier. By varying the strength of the classifier gradient, \citeauthor{dhariwal2021diffusion} can trade off Inception score~\citep{salimans2016improved} and FID score~\citep{heusel2017gans} (or precision and recall) in a manner similar to varying the truncation parameter of BigGAN.

We are interested in whether classifier guidance can be performed without a classifier. Classifier guidance complicates the diffusion model training pipeline because it requires training an extra classifier, and this classifier must be trained on noisy data so it is generally not possible to plug in a pre-trained classifier. Furthermore, because classifier guidance mixes a score estimate with a classifier gradient during sampling, classifier-guided diffusion sampling can be interpreted as attempting to confuse an image classifier with a gradient-based adversarial attack. This raises the question of whether classifier guidance is successful at boosting classifier-based metrics such as FID and Inception score (IS) simply because it is adversarial against such classifiers. Stepping in direction of classifier gradients also bears some resemblance to GAN training, particularly with nonparameteric generators; this also raises the question of whether classifier-guided diffusion models perform well on classifier-based metrics because they are beginning to resemble GANs, which are already known to perform well on such metrics.

To resolve these questions, we present \emph{classifier-free guidance}, our guidance method which avoids any classifier entirely. Rather than sampling in the direction of the gradient of an image classifier, classifier-free guidance instead mixes the score estimates of a conditional diffusion model and a jointly trained unconditional diffusion model. By sweeping over the mixing weight, we attain a FID/IS tradeoff similar to that attained by classifier guidance. Our classifier-free guidance results demonstrate that pure generative diffusion models are capable of synthesizing extremely high fidelity samples possible with other types of generative models.

\section{Background}
\label{sec:background}
We train diffusion models in continuous time~\citep{song2020score,chen2020wavegrad,kingma2021variational}: letting $\bx \sim p(\bx)$ and $\bz = \{\bz_\lambda \,|\, \lambda \in [\lambda_{\mathrm{min}}, \lambda_{\mathrm{max}}]\}$ for hyperparameters $\lambda_{\mathrm{min}} < \lambda_{\mathrm{max}} \in \mathbb{R}$, the forward process $q(\bz|\bx)$ is the variance-preserving Markov process~\citep{sohl2015deep}:
\begin{align}
q(\bz_\lambda|\bx) &= \mathcal{N}(\alpha_\lambda \bx, \sigma_\lambda^2 \bI), \ \text{where}\ \alpha_\lambda^2 = 1/(1+e^{-\lambda}),\ \sigma_\lambda^2 = 1-\alpha_\lambda^2 \\
q(\bz_{\lambda} | \bz_{\lambda'}) &= \mathcal{N}((\alpha_{\lambda}/\alpha_{\lambda'})\bz_{\lambda'}, \sigma_{\lambda|\lambda'}^2\bI), \ \text{where}\ \lambda < \lambda',\ 
\sigma^2_{\lambda|\lambda'} = (1-e^{\lambda-\lambda'})\sigma_\lambda^2 %
\end{align}
We will use the notation $p(\bz)$ (or $p(\bz_\lambda)$) to denote the marginal of $\bz$ (or $\bz_\lambda$) when $\bx \sim p(\bx)$ and $\bz \sim q(\bz|\bx)$. Note that $\lambda = \log \alpha_\lambda^2/\sigma_\lambda^2$, so $\lambda$ can be interpreted as the log signal-to-noise ratio of $\bz_\lambda$, and the forward process runs in the direction of decreasing $\lambda$.

Conditioned on $\bx$, the forward process can be described in reverse by the transitions
$q(\bz_{\lambda'}|\bz_\lambda,\bx) = \mathcal{N}(\tilde\bmu_{\lambda'|\lambda}(\bz_\lambda,\bx), \tilde\sigma^2_{\lambda'|\lambda}\bI)$,
where
\begin{align}
\tilde\bmu_{\lambda'|\lambda}(\bz_\lambda,\bx) = e^{\lambda-\lambda'}(\alpha_{\lambda'}/\alpha_{\lambda})\bz_\lambda + (1-e^{\lambda - \lambda'})\alpha_{\lambda'}\bx,
\quad \tilde\sigma^2_{\lambda'|\lambda} = (1-e^{\lambda-\lambda'})\sigma_{\lambda'}^2
\end{align}
The reverse process generative model starts from $p_\theta(\bz_{\lambda_{\mathrm{min}}}) = \mathcal{N}(\bzero, \bI)$. We specify the transitions:
\begin{align}p_\theta(\bz_{\lambda'}|\bz_{\lambda}) = \mathcal{N}(\tilde\bmu_{\lambda'|\lambda}(\bz_\lambda,\bx_\theta(\bz_\lambda)),  (\tilde\sigma^2_{\lambda'|\lambda})^{1-v} (\sigma^2_{\lambda|\lambda'})^v)\end{align}
During sampling, we apply this transition along an increasing sequence $\lambda_{\mathrm{min}} = \lambda_1 < \cdots < \lambda_T = \lambda_{\mathrm{max}}$ for $T$ timesteps; in other words, we follow the discrete time ancestral sampler of~\citet{sohl2015deep,ho2020denoising}. If the model $\bx_\theta$ is correct, then as $T\rightarrow\infty$, we obtain samples from an SDE whose sample paths are distributed as $p(\bz)$~\citep{song2020score}, and we use $p_\theta(\bz)$ to denote the continuous time model distribution. 
The variance is a log-space interpolation of $\tilde\sigma^2_{\lambda'|\lambda}$ and $\sigma^2_{\lambda|\lambda'}$ as suggested by~\citet{nichol2021improved}; we found it effective to use a constant hyperparameter $v$ rather than learned $\bz_\lambda$-dependent $v$. Note that the variances simplify to $\tilde\sigma^2_{\lambda'|\lambda}$ as $\lambda'\rightarrow\lambda$, so $v$ has an effect only when sampling with non-infinitesimal timesteps as done in practice.

The reverse process mean comes from an estimate $\bx_\theta(\bz_\lambda) \approx \bx$ plugged into $q(\bz_{\lambda'}|\bz_\lambda,\bx)$~\citep{ho2020denoising,kingma2021variational} ($\bx_\theta$ also receives $\lambda$ as input, but we suppress this to keep our notation clean). We parameterize $\bx_\theta$ in terms of $\bepsilon$-prediction~\citep{ho2020denoising}: $\bx_\theta(\bz_\lambda) = (\bz_\lambda - \sigma_\lambda \bepsilon_\theta(\bz_\lambda)) / \alpha_\lambda$, and we train on the objective
\begin{align}
    \Eb{\bepsilon,\lambda}{\|\bepsilon_\theta(\bz_\lambda) - \bepsilon \|^2_2}
\end{align}
where $\bepsilon\sim\mathcal{N}(\bzero,\bI)$, $\bz_\lambda = \alpha_\lambda\bx + \sigma_\lambda\bepsilon$, and $\lambda$ is drawn from a distribution $p(\lambda)$ over $[\lambda_{\mathrm{min}}, \lambda_{\mathrm{max}}]$. This objective is denoising score matching~\citep{vincent2011connection,hyvarinen2005estimation} over multiple noise scales~\citep{song2019generative}, and when $p(\lambda)$ is uniform, the objective is proportional to the variational lower bound on the marginal log likelihood of the latent variable model $\int p_\theta(\bx|\bz) p_\theta(\bz) d\bz$, ignoring the term for the unspecified decoder $p_\theta(\bx|\bz)$ and for the prior at $\bz_{\lambda_\mathrm{min}}$~\citep{kingma2021variational}.

If $p(\lambda)$ is not uniform, the objective can be interpreted as weighted variational lower bound whose weighting can be tuned for sample quality~\citep{ho2020denoising,kingma2021variational}. We use a $p(\lambda)$ inspired by the discrete time cosine noise schedule of~\citet{nichol2021improved}: we sample $\lambda$ via $\lambda = -2\log\tan(au+b)$ for uniformly distributed $u \in [0,1]$, where $b = \arctan(e^{-\lambda_{\mathrm{max}}/2})$ and $a = \arctan(e^{-\lambda_{\mathrm{min}}/2}) - b$. This represents a hyperbolic secant distribution modified to be supported on a bounded interval. For finite timestep generation, we use $\lambda$ values corresponding to uniformly spaced $u \in [0, 1]$, and the final generated sample is $\bx_\theta(\bz_{\lambda_\mathrm{max}})$.

Because the loss for $\bepsilon_\theta(\bz_\lambda)$ is denoising score matching for all $\lambda$,
the score $\bepsilon_\theta(\bz_\lambda)$ learned by our model  estimates the gradient of the log-density of the distribution of our noisy data $\bz_\lambda$, that is $\bepsilon_\theta(\bz_\lambda) \approx -\sigma_\lambda \nabla_{\bz_\lambda}\log p(\bz_\lambda)$; note, however, that because we use unconstrained neural networks to
define $\bepsilon_\theta$, there need not exist any scalar potential whose gradient is $\bepsilon_\theta$. Sampling from the learned diffusion model resembles using Langevin diffusion to sample from a sequence of distributions $p(\bz_\lambda)$ that converges to the conditional distribution $p(\bx)$ of the original data $\bx$.

In the case of conditional generative modeling, the data $\bx$ is drawn jointly with conditioning information~$\bc$, i.e. a class label for class-conditional image generation. The only modification to the model is that the reverse process function approximator receives $\bc$ as input, as in $\bepsilon_\theta(\bz_\lambda, \bc)$.

\section{Guidance}
\label{sec:guidance}

An interesting property of certain generative models, such as GANs and flow-based models, is the ability to perform truncated or low temperature sampling by decreasing the variance or range of noise inputs to the generative model at sampling time. The intended effect is to decrease the diversity of the samples while increasing the quality of each individual sample. Truncation in BigGAN~\citep{brock2018large}, for example, yields a tradeoff curve between FID score and Inception score for low and high amounts of truncation, respectively. Low temperature sampling in Glow~\citep{kingma2018glow} has a similar effect.

Unfortunately, straightforward attempts of implementing truncation or low temperature sampling in diffusion models are ineffective. For example, scaling model scores or decreasing the variance of Gaussian noise in the reverse process cause the diffusion model to generate blurry, low quality samples~\citep{dhariwal2021diffusion}.

\subsection{Classifier guidance}

To obtain a truncation-like effect in diffusion models, \citet{dhariwal2021diffusion} introduce \emph{classifier guidance}, where the diffusion score $\bepsilon_\theta(\bz_\lambda, \bc) \approx -\sigma_\lambda \nabla_{\bz_\lambda}\log p(\bz_\lambda |  \bc)$ is modified to include the gradient of the log likelihood of an auxiliary classifier model $p_{\theta}(\bc | \bz_\lambda)$ as follows:
\[
\tilde{\bepsilon}_\theta(\bz_\lambda, \bc) = \bepsilon_\theta(\bz_\lambda, \bc) - w\sigma_{\lambda}\nabla_{\bz_\lambda}\log p_{\theta}(\bc | \bz_\lambda) \approx -\sigma_{\lambda}\nabla_{\bz_\lambda}[\log p(\bz_\lambda |  \bc) + w \log p_{\theta}(\bc | \bz_\lambda) ],
\]
where $w$ is a parameter that controls the strength of the classifier guidance. This modified score $\tilde{\bepsilon}_\theta(\bz_\lambda, \bc)$ is then used in place of $\bepsilon_\theta(\bz_\lambda, \bc)$ when sampling from the diffusion model, resulting in approximate samples from the distribution
\[\tilde{p}_{\theta}(\bz_\lambda | \bc) \propto p_{\theta}(\bz_\lambda | \bc)p_{\theta}(\bc | \bz_\lambda)^{w}.\]
The effect is that of up-weighting the probability of data for which the classifier $p_{\theta}(\bc | \bz_\lambda)$ assigns high likelihood to the correct label: data that can be classified well scores high on the Inception score of perceptual quality \citep{salimans2016improved}, which rewards generative models for this by design. \citeauthor{dhariwal2021diffusion}
therefore find that by setting $w > 0$ they can improve the Inception score of their diffusion model, at the expense of decreased diversity in their samples.

\Cref{fig:gaussian_guidance} illustrates the effect of numerically solved guidance $\tilde{p}_{\theta}(\bz_\lambda | \bc) \propto p_{\theta}(\bz_\lambda | \bc)p_{\theta}(\bc | \bz_\lambda)^{w}$ on a toy 2D example of three classes, in which the conditional distribution for each class is an isotropic Gaussian. The form of each conditional upon applying guidance is markedly non-Gaussian. As guidance strength is increased, each conditional places probability mass farther away from other classes and towards directions of high confidence given by logistic regression, and most of the mass becomes concentrated in smaller regions. This behavior can be seen as a simplistic manifestation of the Inception score boost and sample diversity decrease that occur when classifier guidance strength is increased in an ImageNet model.

Applying classifier guidance with weight $w+1$ to an unconditional model would theoretically lead to the same result as applying classifier guidance with weight $w$ to a conditional model, because $p_{\theta}(\bz_\lambda | \bc)p_{\theta}(\bc | \bz_\lambda)^{w} \propto p_{\theta}(\bz_\lambda)p_{\theta}(\bc | \bz_\lambda)^{w+1}$; or in terms of scores,
\begin{align*}
\bepsilon_\theta(\bz_\lambda) - (w+1)\sigma_{\lambda}\nabla_{\bz_\lambda}\log p_{\theta}(\bc | \bz_\lambda) &\approx -\sigma_{\lambda}\nabla_{\bz_\lambda}[\log p(\bz_\lambda) + (w+1) \log p_{\theta}(\bc | \bz_\lambda) ] \\
&= -\sigma_{\lambda}\nabla_{\bz_\lambda}[\log p(\bz_\lambda|\bc) + w\log p_{\theta}(\bc | \bz_\lambda) ],
\end{align*}
but interestingly, \citeauthor{dhariwal2021diffusion} obtain their best results when applying classifier guidance to an already class-conditional model, as opposed to applying guidance to an unconditional model. For this reason, we will stay in the setup of guiding an already conditional model.

\subsection{Classifier-free guidance}

While classifier guidance successfully trades off IS and FID as expected from truncation or low temperature sampling, it is nonetheless reliant on gradients from an image classifier and we seek to eliminate the classifier for the reasons stated in~\cref{sec:introduction}. Here, we describe \emph{classifier-free guidance}, which achieves the same effect without such gradients. Classifier-free guidance is an alternative method of modifying $\bepsilon_\theta(\bz_\lambda, \bc)$ to have the same effect as classifier guidance, but without a classifier. \cref{alg:training,alg:sampling} describe training and sampling with classifier-free guidance in detail.

\begin{algorithm}[tb]
  \caption{Joint training a diffusion model with classifier-free guidance} \label{alg:training}
  \begin{algorithmic}[1]
    \Require $p_\mathrm{uncond}$: probability of unconditional training
    \Repeat
      \State $(\bx,\bc) \sim p(\bx,\bc)$ \Comment{Sample data with conditioning from the dataset}
      \State $\bc \gets \varnothing$ with probability $p_\mathrm{uncond}$ \Comment{Randomly discard conditioning to train unconditionally}
      \State $\lambda \sim p(\lambda)$ \Comment{Sample log SNR value}
      \State $\bepsilon\sim\mathcal{N}(\bzero,\bI)$
      \State $\bz_\lambda = \alpha_\lambda\bx + \sigma_\lambda \bepsilon$ \Comment{Corrupt data to the sampled log SNR value}
      \State Take gradient step on $\grad_\theta \left\| \bepsilon_\theta(\bz_\lambda,\bc) - \bepsilon \right\|^2$ \Comment{Optimization of denoising model}
    \Until{converged}
  \end{algorithmic}
\end{algorithm}

\begin{algorithm}[tb]
  \caption{Conditional sampling with classifier-free guidance} \label{alg:sampling}
  \begin{algorithmic}[1]
    \Require $w$: guidance strength
    \Require $\bc$: conditioning information for conditional sampling
    \Require $\lambda_1, \dotsc, \lambda_T$: increasing log SNR sequence with  $\lambda_1=\lambda_{\mathrm{min}}$,  $\lambda_T=\lambda_{\mathrm{max}}$
    \State $\bz_{1} \sim \mathcal{N}(\bzero, \bI)$
    \For{$t=1, \dotsc, T$}

      $\!\!\triangleright$ Form the classifier-free guided score at log SNR $\lambda_t$
      \State $\tilde{\bepsilon}_t = (1+w)\bepsilon_\theta(\bz_{t}, \bc) - w\bepsilon_{\theta}(\bz_{t})$ 

      $\!\!\triangleright$ Sampling step (could be replaced by another sampler, e.g. DDIM)
      \State $\tilde\bx_t = (\bz_{t}-\sigma_{\lambda_t}\tilde\bepsilon_t)/\alpha_{\lambda_t}$
      \State $\bz_{t+1} \sim \mathcal{N}(\tilde\bmu_{\lambda_{t+1}|\lambda_t}(\bz_{t},\tilde\bx_t),  (\tilde\sigma^2_{\lambda_{t+1}|\lambda_t})^{1-v} (\sigma^2_{\lambda_t|\lambda_{t+1}})^v)$ if $t<T$ else $\bz_{t+1}=\tilde\bx_t$
    \EndFor
    \State \textbf{return} $\bz_{T+1}$
  \end{algorithmic}
\end{algorithm}

Instead of training a separate classifier model, we choose to train an unconditional denoising diffusion model $p_{\theta}(\bz)$ parameterized through a score estimator $\bepsilon_{\theta}(\bz_\lambda)$ together with the conditional model $p_{\theta}(\bz | \bc)$ parameterized through $\bepsilon_{\theta}(\bz_\lambda, \bc)$. We use a single neural network to parameterize both models, where for the unconditional model we can simply input a null token $\varnothing$ for the class identifier $\bc$ when predicting the score, i.e.\ $\bepsilon_{\theta}(\bz_\lambda) = \bepsilon_{\theta}(\bz_\lambda, \bc = \varnothing)$. We jointly train the unconditional and conditional models simply by randomly setting $\bc$ to the unconditional class identifier $\varnothing$ with some probability $p_\mathrm{uncond}$, set as a hyperparameter. (It would certainly be possible to train separate models instead of jointly training them together, but we  choose joint training because it is extremely simple to implement, does not complicate the training pipeline, and does not increase the total number of parameters.)
We then perform sampling using the following linear combination of the conditional and unconditional score estimates:
\begin{align}
    \tilde{\bepsilon}_\theta(\bz_\lambda, \bc) = (1+w)\bepsilon_\theta(\bz_\lambda, \bc) - w\bepsilon_{\theta}(\bz_\lambda) \label{eq:classifier_free_score}
\end{align}
\cref{eq:classifier_free_score} has no classifier gradient present, so taking a step in the $\tilde\bepsilon_\theta$ direction cannot be interpreted as a gradient-based adversarial attack on an image classifier. Furthermore, $\tilde\bepsilon_\theta$ is constructed from score estimates that are non-conservative vector fields due to the use of unconstrained neural networks, so there in general cannot exist a scalar potential such as a classifier log likelihood for which $\tilde\bepsilon_\theta$ is the classifier-guided score. 

Despite the fact that there in general may not exist a classifier for which  \cref{eq:classifier_free_score} is the classifier-guided score, it is in fact inspired by the gradient of an implicit classifier $p^{i}(\bc | \bz_\lambda) \propto p(\bz_\lambda | \bc)/p(\bz_\lambda)$. If we had access to exact scores $\bepsilon^*(\bz_\lambda, \bc)$ and $\bepsilon^*(\bz_\lambda)$ (of $p(\bz_\lambda|\bc)$ and $p(\bz_\lambda)$, respectively), then the gradient of this implicit classifier would be $\nabla_{\bz_\lambda} \log p^{i}(\bc | \bz_\lambda) = -\frac{1}{\sigma_{\lambda}}[\bepsilon^*(\bz_\lambda, \bc) - \bepsilon^*(\bz_\lambda)]$, and classifier guidance with this implicit classifier would modify the score estimate into $\tilde{\bepsilon}^*(\bz_\lambda, \bc) = (1+w)\bepsilon^*(\bz_\lambda, \bc) - w\bepsilon^*(\bz_\lambda)$. Note the resemblance to \cref{eq:classifier_free_score}, but also note that $\tilde{\bepsilon}^*(\bz_\lambda,\bc)$ differs fundamentally from  $\tilde{\bepsilon}_\theta(\bz_\lambda,\bc)$. The former is constructed from the scaled classifier gradient $\bepsilon^*(\bz_\lambda, \bc) - \bepsilon^*(\bz_\lambda)$; the latter is constructed from the estimate $\bepsilon_\theta(\bz_\lambda, \bc) - \bepsilon_\theta(\bz_\lambda)$, and this expression is not in general the (scaled) gradient of any classifier, again because the score estimates are the outputs of unconstrained neural networks.

It is not obvious a priori that inverting a generative model using Bayes' rule yields a good classifier that provides a useful guidance signal. For example, \cite{grandvalet2004semi} find that discriminative models generally outperform implicit classifiers derived from generative models, even in artificial cases where the specification of those generative models exactly matches the data distribution. In cases such as ours, where we expect the model to be misspecified, classifiers derived by Bayes' rule can be inconsistent~\citep{grunwald2007suboptimal} and we lose all guarantees on their performance. Nevertheless, in \cref{sec:experiments}, we show empirically that classifier-free guidance is able to trade off FID and IS in the same way as classifier guidance. In \cref{sec:discussion} we discuss the implications of classifier-free guidance in relation to classifier guidance.

\section{Experiments}
\label{sec:experiments}

\begin{figure}[tb] \centering
\includegraphics[width=\linewidth]{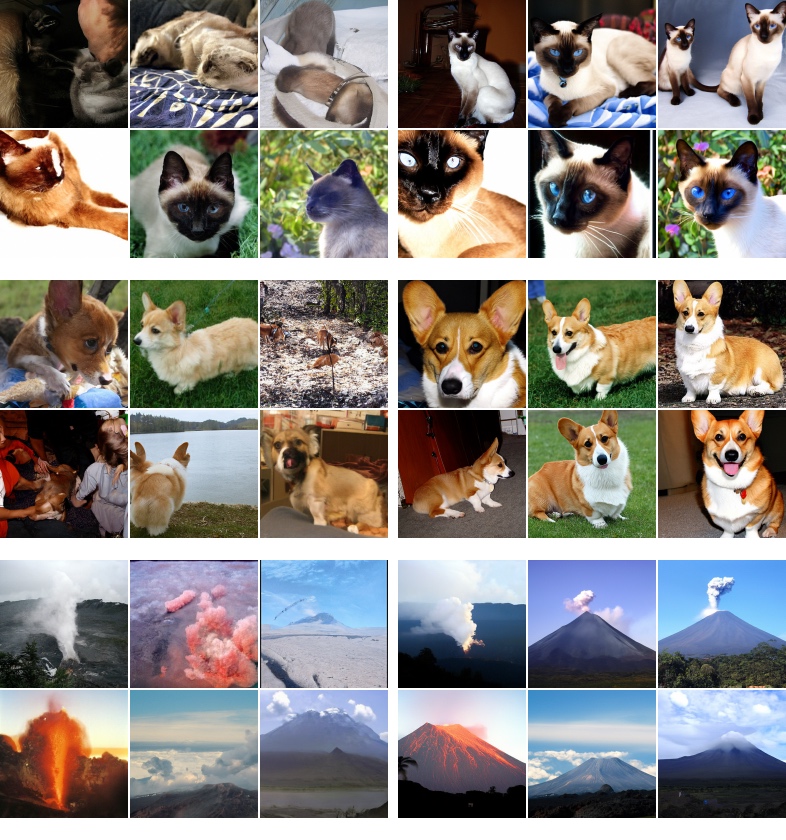}
\vspace{-2em}
\caption{Classifier-free guidance on 128x128 ImageNet. Left: non-guided samples, right: classifier-free guided samples with $w=3.0$. Interestingly, strongly guided samples such as these display saturated colors. See \cref{fig:i128_guidance_more} for more.}
\label{fig:i128_guidance}
\end{figure}

We train diffusion models with classifier-free guidance on area-downsampled class-conditional ImageNet~\citep{russakovsky2015imagenet}, the standard setting for studying tradeoffs between FID and Inception scores starting from the BigGAN paper~\citep{brock2018large}.

The purpose of our experiments is to serve as a proof of concept to demonstrate that classifier-free guidance is able to attain a FID/IS tradeoff similar to classifier guidance and to understand the behavior of classifier-free guidance, not necessarily to push sample quality metrics to state of the art on these benchmarks. For this purpose, we use the same model architectures and hyperparameters as the guided diffusion models of \citet{dhariwal2021diffusion} (apart from continuous time training as specified in \cref{sec:background}); those hyperparameter settings were tuned for classifier guidance and hence may be suboptimal for classifier-free guidance. Furthermore, since we amortize the conditional and unconditional models into the same architecture without an extra classifier, we in fact are using less model capacity than previous work. Nevertheless, our classifier-free guided models still produce competitive sample quality metrics and sometimes outperform prior work, as can be seen in the following sections.

\subsection{Varying the classifier-free guidance strength}
\begin{figure}[!htb] \centering
\begin{tikzpicture}
\begin{axis}[xlabel=IS,ylabel=FID,legend pos=north west,width=10cm,height=6cm]
\addplot[color=blue,mark=x] table[x=is1,y=fid1] {\loadedtable};
\addplot[color=red,mark=x] table[x=is2,y=fid2] {\loadedtable};
\addplot[color=black,mark=x] table[x=is5,y=fid5] {\loadedtable};
\legend{$p_\mathrm{uncond}=0.1$,$p_\mathrm{uncond}=0.2$,$p_\mathrm{uncond}=0.5$}
\end{axis}
\end{tikzpicture}
\caption{IS/FID curves over guidance strengths for ImageNet 64x64 models. Each curve represents a model with unconditional training probability $p_\mathrm{uncond}$. Accompanies~\cref{table:i64}.}
\label{fig:i64_fid_is_plot}
\end{figure}
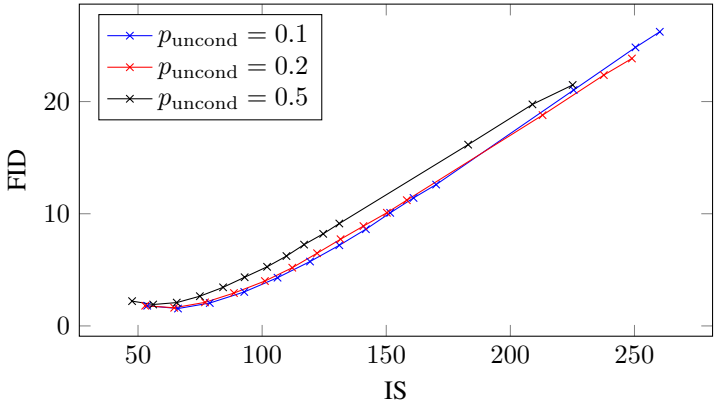

Here we experimentally verify the main claim of this paper: that classifier-free guidance is able to trade off IS and FID in a manner like classifier guidance or GAN truncation.
We apply our proposed classifier-free guidance to $64\times64$ and $128\times128$ class-conditional ImageNet generation. 
In \cref{table:i64} and \cref{fig:i64_fid_is_plot}, we show sample quality effects of sweeping over the guidance strength $w$ on our $64\times64$ ImageNet models; \cref{table:i128} and \cref{fig:i128_fid_is_plot} show the same for our $128\times128$ models. 
We consider $w \in \{0, 0.1, 0.2, \ldots, 4\}$ and calculate FID and Inception Scores with 50000 samples for each value following the procedures of \citet{heusel2017gans} and \citet{salimans2016improved}. All models used log SNR endpoints $\lambda_\mathrm{min}=-20$ and $\lambda_\mathrm{max}=20$. The $64\times64$ models used sampler noise interpolation coefficient $v=0.3$ and were trained for 400 thousand steps; the $128\times128$ models used $v=0.2$ and were trained for 2.7 million steps.

We obtain the best FID results with a small amount of guidance ($w = 0.1$ or $w=0.3$, depending on the dataset) and the best IS result with strong guidance ($w \geq 4$). Between these two extremes we see a clear trade-off between these two metrics of perceptual quality, with FID monotonically decreasing and IS monotonically increasing with $w$. Our results compare favorably to \citet{dhariwal2021diffusion} and \citet{ho2021cascaded}, and in fact our $128\times128$ results are the state of the art in the literature. At $w=0.3$, our model's FID score on $128\times128$ ImageNet outperforms the classifier-guided ADM-G, and at $w=4.0$, our model outperforms BigGAN-deep at both FID and IS when BigGAN-deep is evaluated its best-IS truncation level.

\Cref{fig:dog_guidance,fig:i64_samples,fig:i128_guidance,fig:i128_samples,fig:i128_guidance_more} show randomly generated samples from our model for different levels of guidance: here we clearly see that increasing classifier-free guidance strength has the expected effect of decreasing sample variety and increasing individual sample fidelity.

\begin{table}[!htb]
\centering\small
\begin{tabular}{c|c|c}
\toprule
Model & FID ($\downarrow$) & IS ($\uparrow$) \\
\midrule
ADM~\citep{dhariwal2021diffusion} & 2.07 & - \\
CDM~\citep{ho2021cascaded} & \textbf{1.48} & 67.95 \\
\midrule
Ours & \multicolumn{2}{c}{$p_\mathrm{uncond}=0.1/0.2/0.5$} \\
$w=0.0$  & 1.8 / 1.8 / 2.21    &    53.71 / 52.9 / 47.61 \\
$w=0.1$  & 1.55 / 1.62 / 1.91    &    66.11 / 64.58 / 56.1 \\
$w=0.2$  & 2.04 / 2.1 / 2.08    &    78.91 / 76.99 / 65.6 \\
$w=0.3$  & 3.03 / 2.93 / 2.65    &    92.8 / 88.64 / 74.92 \\
$w=0.4$  & 4.3 / 4 / 3.44    &    106.2 / 101.11 / 84.27 \\
$w=0.5$  & 5.74 / 5.19 / 4.34    &    119.3 / 112.15 / 92.95 \\
$w=0.6$  & 7.19 / 6.48 / 5.27    &    131.1 / 122.13 / 102 \\
$w=0.7$  & 8.62 / 7.73 / 6.23    &    141.8 / 131.6 / 109.8 \\
$w=0.8$  & 10.08 / 8.9 / 7.25    &    151.6 / 140.82 / 116.9 \\
$w=0.9$  & 11.41 / 10.09 / 8.21    &    161 / 150.26 / 124.6 \\
$w=1.0$  & 12.6 / 11.21 / 9.13    &    170.1 / 158.29 / 131.1 \\
$w=2.0$  & 21.03 / 18.79 / 16.16    &    225.5 / 212.98 / 183 \\
$w=3.0$  & 24.83 / 22.36 / 19.75    &    250.4 / 237.65 / 208.9 \\
$w=4.0$  & 26.22 / 23.84 / 21.48    &    \textbf{260.2} / 248.97 / 225.1 \\
\bottomrule
\end{tabular}
\caption{ImageNet 64x64 results ($w=0.0$ refers to non-guided models).}
\label{table:i64}
\end{table}

\pgfplotstableread{
fid1 fid2 fid5 is1 is2 is5
1.8	1.8	2.21		53.71	52.9	47.61
1.55	1.62	1.91		66.11	64.58	56.1
2.04	2.1	2.08		78.91	76.99	65.6
3.03	2.93	2.65		92.8	88.64	74.92
4.3	4	3.44		106.2	101.11	84.27
5.74	5.19	4.34		119.3	112.15	92.95
7.19	6.48	5.27		131.1	122.13	102
8.62	7.73	6.23		141.8	131.6	109.8
10.08	8.9	7.25		151.6	140.82	116.9
11.41	10.09	8.21		161	150.26	124.6
12.6	11.21	9.13		170.1	158.29	131.1
21.03	18.79	16.16		225.5	212.98	183
24.83	22.36	19.75		250.4	237.65	208.9
26.22	23.84	21.48		260.2	248.97	225.1
}\loadedtable

\subsection{Varying the unconditional training probability}

The main hyperparameter of classifier-free guidance at training time is $p_\mathrm{uncond}$, the probability of training on unconditional generation during joint training of the conditional and unconditional diffusion models. Here, we study the effect of training models on varying $p_\mathrm{uncond}$ on $64\times64$ ImageNet.

\cref{table:i64} and \cref{fig:i64_fid_is_plot} show the effects of $p_\mathrm{uncond}$ on sample quality. We trained models with~$p_\mathrm{uncond}\in\{0.1,0.2,0.5\}$, all for 400 thousand training steps, and evaluated sample quality across various guidance strengths. We find $p_\mathrm{uncond}=0.5$ consistently performs worse than $p_\mathrm{uncond}\in\{0.1,0.2\}$ across the entire IS/FID frontier; $p_\mathrm{uncond}\in\{0.1,0.2\}$ perform about equally as well as each other.

Based on these findings, we conclude that only a relatively small portion of the model capacity of the diffusion model needs to be dedicated to the unconditional generation task in order to produce classifier-free guided scores effective for sample quality. Interestingly, for classifier guidance, \citeauthor{dhariwal2021diffusion} report that relatively small classifiers with little capacity are sufficient for effective classifier guided sampling, mirroring this phenomenon that we found with classifier-free guided models.

\subsection{Varying the number of sampling steps}

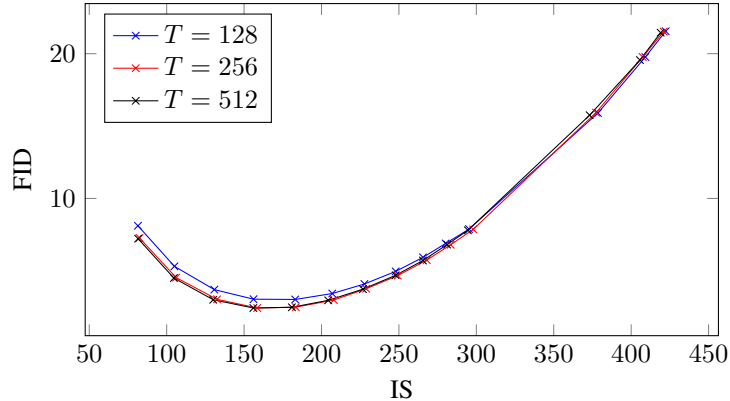
\begin{figure}[!htb] \centering
\begin{tikzpicture}
\begin{axis}[xlabel=IS,ylabel=FID,legend pos=north west,width=10cm,height=6cm]
\addplot[color=blue,mark=x] table[x=is1,y=fid1] {\loadedtable};
\addplot[color=red,mark=x] table[x=is2,y=fid2] {\loadedtable};
\addplot[color=black,mark=x] table[x=is5,y=fid5] {\loadedtable};
\legend{$T=128$,$T=256$,$T=512$}
\end{axis}
\end{tikzpicture}
\caption{IS/FID curves over guidance strengths for ImageNet 128x128 models. Each curve represents sampling with a different number of timesteps $T$. Accompanies~\cref{table:i128}.}
\label{fig:i128_fid_is_plot}
\end{figure}

Since the number of sampling steps $T$ is known to have a major impact on the sample quality of a diffusion model, here we study the effect of varying $T$ on our $128\times128$ ImageNet model. \cref{table:i128} and \cref{fig:i128_fid_is_plot} show the effect of varying $T\in\{128,256,1024\}$ over a range of guidance strengths. As expected, sample quality improves when $T$ is increased, and for this model $T=256$ attains a good balance between sample quality and sampling speed.

Note that $T=256$ is approximately the same number of sampling steps used by ADM-G ~\citep{dhariwal2021diffusion}, which is outperformed by our model. However, it is important to note that each sampling step for our method requires evaluating the denoising model twice, once for the conditional $\bepsilon_\theta(\bz_\lambda,\bc)$ and once for the unconditional $\bepsilon_\theta(\bz_\lambda)$. Because we used the same model architecture as ADM-G, the fair comparison in terms of sampling speed would be our $T=128$ setting, which underperforms compared to ADM-G in terms of FID score.

\begin{table}[!htb]
\centering\small
\begin{tabular}{c|c|c}
\toprule
Model & FID ($\downarrow$) & IS ($\uparrow$) \\
\midrule
BigGAN-deep, max IS~\citep{brock2018large} & 25 & 253 \\
BigGAN-deep~\citep{brock2018large} & 5.7 & 124.5 \\
CDM~\citep{ho2021cascaded} & 3.52 & 128.8 \\
LOGAN~\citep{wu2019logan} & 3.36 & 148.2 \\
ADM-G~\citep{dhariwal2021diffusion} & 2.97 & - \\
\midrule
Ours & \multicolumn{2}{c}{$T=128/256/1024$} \\
$w=0.0$  & 8.11 / 7.27 / 7.22    &    81.46 / 82.45 / 81.54 \\
$w=0.1$  & 5.31 / 4.53 / 4.5    &    105.01 / 106.12 / 104.67 \\
$w=0.2$  & 3.7 / 3.03 / 3    &    130.79 / 132.54 / 130.09 \\
$w=0.3$  & 3.04 / \textbf{2.43} / \textbf{2.43}    &    156.09 / 158.47 / 156 \\
$w=0.4$  & 3.02 / 2.49 / 2.48    &    183.01 / 183.41 / 180.88 \\
$w=0.5$  & 3.43 / 2.98 / 2.96    &    206.94 / 207.98 / 204.31 \\
$w=0.6$  & 4.09 / 3.76 / 3.73    &    227.72 / 228.83 / 226.76 \\
$w=0.7$  & 4.96 / 4.67 / 4.69    &    247.92 / 249.25 / 247.89 \\
$w=0.8$  & 5.93 / 5.74 / 5.71    &    265.54 / 267.99 / 265.52 \\
$w=0.9$  & 6.89 / 6.8 / 6.81    &    280.19 / 283.41 / 281.14 \\
$w=1.0$  & 7.88 / 7.86 / 7.8    &    295.29 / 297.98 / 294.56 \\
$w=2.0$  & 15.9 / 15.93 / 15.75    &    378.56 / 377.37 / 373.18 \\
$w=3.0$  & 19.77 / 19.77 / 19.56    &    409.16 / 407.44 / 405.68 \\
$w=4.0$  & 21.55 / 21.53 / 21.45    &    \textbf{422.29} / 421.03 / 419.06 \\
\bottomrule
\end{tabular}
\caption{ImageNet 128x128 results ($w=0.0$ refers to non-guided models).}
\label{table:i128}
\end{table}

\pgfplotstableread{
fid1 fid2 fid5 is1 is2 is5
8.11	7.27	7.22		81.46	82.45	81.54
5.31	4.53	4.5		105.01	106.12	104.67
3.7	3.03	3		130.79	132.54	130.09
3.04	2.43	2.43		156.09	158.47	156
3.02	2.49	2.48		183.01	183.41	180.88
3.43	2.98	2.96		206.94	207.98	204.31
4.09	3.76	3.73		227.72	228.83	226.76
4.96	4.67	4.69		247.92	249.25	247.89
5.93	5.74	5.71		265.54	267.99	265.52
6.89	6.8	6.81		280.19	283.41	281.14
7.88	7.86	7.8		295.29	297.98	294.56
15.9	15.93	15.75		378.56	377.37	373.18
19.77	19.77	19.56		409.16	407.44	405.68
21.55	21.53	21.45		422.29	421.03	419.06
}\loadedtable

\section{Discussion}
\label{sec:discussion}

The most practical advantage of our classifier-free guidance method is its extreme simplicity: it is only a one-line change of code during training---to randomly drop out the conditioning---and during sampling---to mix the conditional and unconditional score estimates. Classifier guidance, by contrast, complicates the training pipeline since it requires training an extra classifier. This classifier must be trained on noisy $\bz_\lambda$, so it is not possible to plug in a standard pre-trained classifier.

Since classifier-free guidance is able to trade off IS and FID like classifier guidance without needing an extra trained classifier, we have demonstrated that guidance can be performed with a pure generative model. 
Furthermore, our diffusion models are parameterized by unconstrained neural networks and therefore their score estimates do not necessarily form conservative vector fields, unlike classifier gradients~\citep{salimans2021should}. Therefore, our  classifier-free guided sampler follows step directions that do not resemble classifier gradients at all and thus cannot be interpreted as a gradient-based adversarial attack on a classifier, and hence our results show that boosting the classifier-based IS and FID metrics can be accomplished with pure generative models with a sampling procedure that is not adversarial against image classifiers using classifier gradients.

We also have arrived at an intuitive explanation for how guidance works: it decreases the unconditional likelihood of the sample while increasing the conditional likelihood. %
Classifier-free guidance accomplishes this by decreasing the unconditional likelihood with a \emph{negative} score term, which to our knowledge has not yet been explored and may find uses in other applications.

Classifier-free guidance as presented here relies on training an unconditional model, but in some cases this can be avoided. If the class distribution is known and there are only a few classes, we can use the fact that $\sum_\bc p(\bx|\bc) p(\bc) = p(\bx)$ to obtain an unconditional score from conditional scores without explicitly training for the unconditional score. Of course, this would require as many forward passes as there are possible values of $\bc$ and would be inefficient for high dimensional conditioning.

A potential disadvantage of classifier-free guidance is sampling speed. Generally, classifiers can be smaller and faster than generative models, so classifier guided sampling may be faster than classifier-free guidance because the latter needs to run two forward passes of the diffusion model, one for conditional score and another for the unconditional score. The necessity to run multiple passes of the diffusion model might be mitigated by changing the architecture to inject conditioning late in the network, but we leave this exploration for future work.

Finally, any guidance method that increases sample fidelity at the expense of diversity must face the question of whether decreased diversity is acceptable. There may be negative impacts in deployed models, since sample diversity is important to maintain in applications where certain parts of the data are underrepresented in the context of the rest of the data. It would be an interesting avenue of future work to try to boost sample quality while maintaining sample diversity.

\section{Conclusion}

We have presented classifier-free guidance, a method to increase sample quality while decreasing sample diversity in diffusion models. Classifier-free guidance can be thought of as classifier guidance without a classifier, and our results showing the effectiveness of classifier-free guidance confirm that pure generative diffusion models are capable of maximizing classifier-based sample quality metrics while entirely avoiding classifier gradients. We look forward to further explorations of classifier-free guidance in a wider variety of settings and data modalities.

\section{Acknowledgements}
We thank Ben Poole and Mohammad Norouzi for discussions.

\FloatBarrier

\bibliography{main}

\begin{thebibliography}{23}
\providecommand{\natexlab}[1]{#1}
\providecommand{\url}[1]{\texttt{#1}}
\expandafter\ifx\csname urlstyle\endcsname\relax
  \providecommand{\doi}[1]{doi: #1}\else
  \providecommand{\doi}{doi: \begingroup \urlstyle{rm}\Url}\fi

\bibitem[Brock et~al.(2019)Brock, Donahue, and Simonyan]{brock2018large}
Andrew Brock, Jeff Donahue, and Karen Simonyan.
\newblock Large scale {GAN} training for high fidelity natural image synthesis.
\newblock In \emph{International Conference on Learning Representations}, 2019.

\bibitem[Chen et~al.(2021)Chen, Zhang, Zen, Weiss, Norouzi, and
  Chan]{chen2020wavegrad}
Nanxin Chen, Yu~Zhang, Heiga Zen, Ron~J Weiss, Mohammad Norouzi, and William
  Chan.
\newblock {WaveGrad}: Estimating gradients for waveform generation.
\newblock \emph{{International Conference on Learning Representations}}, 2021.

\bibitem[Dhariwal \& Nichol(2021)Dhariwal and Nichol]{dhariwal2021diffusion}
Prafulla Dhariwal and Alex Nichol.
\newblock Diffusion models beat {GAN}s on image synthesis.
\newblock \emph{arXiv preprint arXiv:2105.05233}, 2021.

\bibitem[Grandvalet \& Bengio(2004)Grandvalet and Bengio]{grandvalet2004semi}
Yves Grandvalet and Yoshua Bengio.
\newblock Semi-supervised learning by entropy minimization.
\newblock In \emph{Proceedings of the 17th International Conference on Neural
  Information Processing Systems}, pp.\  529--536, 2004.

\bibitem[Gr{\"u}nwald \& Langford(2007)Gr{\"u}nwald and
  Langford]{grunwald2007suboptimal}
Peter Gr{\"u}nwald and John Langford.
\newblock Suboptimal behavior of bayes and mdl in classification under
  misspecification.
\newblock \emph{Machine Learning}, 66\penalty0 (2-3):\penalty0 119--149, 2007.

\bibitem[Heusel et~al.(2017)Heusel, Ramsauer, Unterthiner, Nessler, and
  Hochreiter]{heusel2017gans}
Martin Heusel, Hubert Ramsauer, Thomas Unterthiner, Bernhard Nessler, and Sepp
  Hochreiter.
\newblock {GANs} trained by a two time-scale update rule converge to a local
  {Nash} equilibrium.
\newblock In \emph{Advances in Neural Information Processing Systems}, pp.\
  6626--6637, 2017.

\bibitem[Ho et~al.(2020)Ho, Jain, and Abbeel]{ho2020denoising}
Jonathan Ho, Ajay Jain, and Pieter Abbeel.
\newblock Denoising diffusion probabilistic models.
\newblock In \emph{Advances in Neural Information Processing Systems}, pp.\
  6840--6851, 2020.

\bibitem[Ho et~al.(2021)Ho, Saharia, Chan, Fleet, Norouzi, and
  Salimans]{ho2021cascaded}
Jonathan Ho, Chitwan Saharia, William Chan, David~J Fleet, Mohammad Norouzi,
  and Tim Salimans.
\newblock Cascaded diffusion models for high fidelity image generation.
\newblock \emph{arXiv preprint arXiv:2106.15282}, 2021.

\bibitem[Hyv{\"a}rinen \& Dayan(2005)Hyv{\"a}rinen and
  Dayan]{hyvarinen2005estimation}
Aapo Hyv{\"a}rinen and Peter Dayan.
\newblock Estimation of non-normalized statistical models by score matching.
\newblock \emph{Journal of Machine Learning Research}, 6\penalty0 (4), 2005.

\bibitem[Kingma \& Dhariwal(2018)Kingma and Dhariwal]{kingma2018glow}
Diederik~P Kingma and Prafulla Dhariwal.
\newblock Glow: Generative flow with invertible 1x1 convolutions.
\newblock In \emph{Advances in Neural Information Processing Systems}, pp.\
  10215--10224, 2018.

\bibitem[Kingma et~al.(2021)Kingma, Salimans, Poole, and
  Ho]{kingma2021variational}
Diederik~P Kingma, Tim Salimans, Ben Poole, and Jonathan Ho.
\newblock Variational diffusion models.
\newblock \emph{arXiv preprint arXiv:2107.00630}, 2021.

\bibitem[Kong et~al.(2021)Kong, Ping, Huang, Zhao, and
  Catanzaro]{kong2020diffwave}
Zhifeng Kong, Wei Ping, Jiaji Huang, Kexin Zhao, and Bryan Catanzaro.
\newblock {DiffWave: A Versatile Diffusion Model for Audio Synthesis}.
\newblock \emph{{International Conference on Learning Representations}}, 2021.

\bibitem[Nichol \& Dhariwal(2021)Nichol and Dhariwal]{nichol2021improved}
Alex Nichol and Prafulla Dhariwal.
\newblock Improved denoising diffusion probabilistic models.
\newblock \emph{{International Conference on Machine Learning}}, 2021.

\bibitem[Razavi et~al.(2019)Razavi, van~den Oord, and
  Vinyals]{razavi2019generating}
Ali Razavi, Aaron van~den Oord, and Oriol Vinyals.
\newblock Generating diverse high-fidelity images with {VQ-VAE-2}.
\newblock In \emph{Advances in Neural Information Processing Systems}, pp.\
  14837--14847, 2019.

\bibitem[Russakovsky et~al.(2015)Russakovsky, Deng, Su, Krause, Satheesh, Ma,
  Huang, Karpathy, Khosla, Bernstein, et~al.]{russakovsky2015imagenet}
Olga Russakovsky, Jia Deng, Hao Su, Jonathan Krause, Sanjeev Satheesh, Sean Ma,
  Zhiheng Huang, Andrej Karpathy, Aditya Khosla, Michael Bernstein, et~al.
\newblock {ImageNet} large scale visual recognition challenge.
\newblock \emph{International Journal of Computer Vision}, 115\penalty0
  (3):\penalty0 211--252, 2015.

\bibitem[Salimans \& Ho(2021)Salimans and Ho]{salimans2021should}
Tim Salimans and Jonathan Ho.
\newblock Should {EBM}s model the energy or the score?
\newblock In \emph{Energy Based Models Workshop-ICLR 2021}, 2021.

\bibitem[Salimans et~al.(2016)Salimans, Goodfellow, Zaremba, Cheung, Radford,
  and Chen]{salimans2016improved}
Tim Salimans, Ian Goodfellow, Wojciech Zaremba, Vicki Cheung, Alec Radford, and
  Xi~Chen.
\newblock Improved techniques for training {GAN}s.
\newblock In \emph{Advances in Neural Information Processing Systems}, pp.\
  2234--2242, 2016.

\bibitem[Sohl-Dickstein et~al.(2015)Sohl-Dickstein, Weiss, Maheswaranathan, and
  Ganguli]{sohl2015deep}
Jascha Sohl-Dickstein, Eric Weiss, Niru Maheswaranathan, and Surya Ganguli.
\newblock Deep unsupervised learning using nonequilibrium thermodynamics.
\newblock In \emph{International Conference on Machine Learning}, pp.\
  2256--2265, 2015.

\bibitem[Song \& Ermon(2019)Song and Ermon]{song2019generative}
Yang Song and Stefano Ermon.
\newblock Generative modeling by estimating gradients of the data distribution.
\newblock In \emph{Advances in Neural Information Processing Systems}, pp.\
  11895--11907, 2019.

\bibitem[Song et~al.(2021{\natexlab{a}})Song, Durkan, Murray, and
  Ermon]{song2021maximum}
Yang Song, Conor Durkan, Iain Murray, and Stefano Ermon.
\newblock Maximum likelihood training of score-based diffusion models.
\newblock \emph{arXiv e-prints}, pp.\  arXiv--2101, 2021{\natexlab{a}}.

\bibitem[Song et~al.(2021{\natexlab{b}})Song, Sohl-Dickstein, Kingma, Kumar,
  Ermon, and Poole]{song2020score}
Yang Song, Jascha Sohl-Dickstein, Diederik~P Kingma, Abhishek Kumar, Stefano
  Ermon, and Ben Poole.
\newblock Score-based generative modeling through stochastic differential
  equations.
\newblock \emph{{International Conference on Learning Representations}},
  2021{\natexlab{b}}.

\bibitem[Vincent(2011)]{vincent2011connection}
Pascal Vincent.
\newblock A connection between score matching and denoising autoencoders.
\newblock \emph{Neural Computation}, 23\penalty0 (7):\penalty0 1661--1674,
  2011.

\bibitem[Wu et~al.(2019)Wu, Donahue, Balduzzi, Simonyan, and
  Lillicrap]{wu2019logan}
Yan Wu, Jeff Donahue, David Balduzzi, Karen Simonyan, and Timothy Lillicrap.
\newblock {LOGAN}: Latent optimisation for generative adversarial networks.
\newblock \emph{arXiv preprint arXiv:1912.00953}, 2019.

\end{thebibliography}
\bibliographystyle{iclr2022_conference}

\newpage
\appendix
\section{Samples}

\begin{figure}[H]
     \centering
     \begin{subfigure}[b]{\textwidth}
         \centering
         \includegraphics[width=0.9\linewidth]{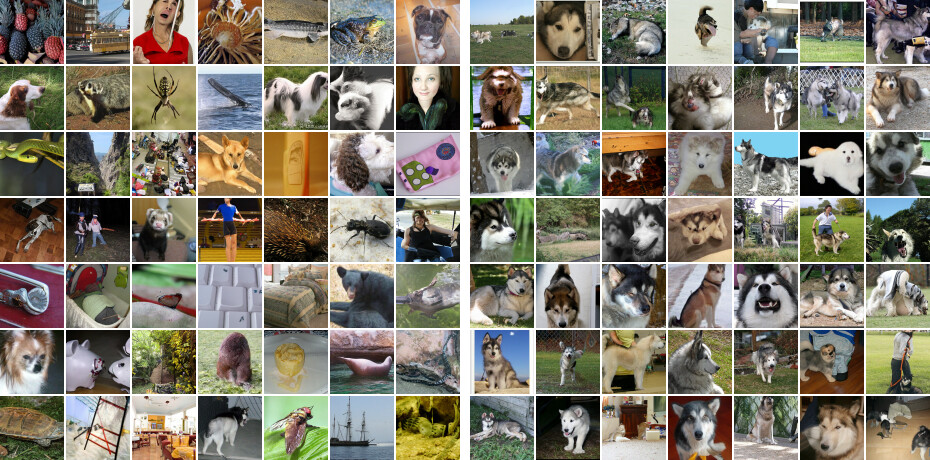}
         \caption{Non-guided conditional sampling: FID=1.80, IS=53.71}
     \end{subfigure}
     \begin{subfigure}[b]{\textwidth}
         \centering
         \includegraphics[width=0.9\linewidth]{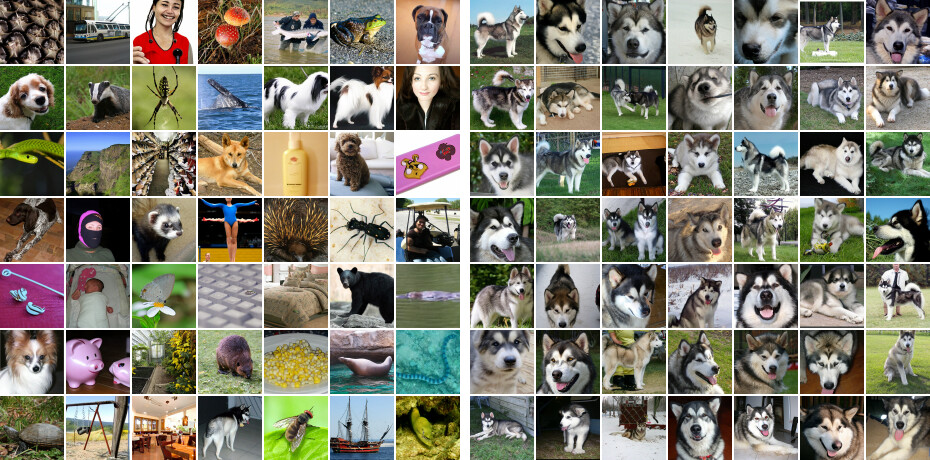}
         \caption{Classifier-free guidance with $w=1.0$: FID=12.6, IS=170.1}
     \end{subfigure}
     \begin{subfigure}[b]{\textwidth}
         \centering
         \includegraphics[width=0.9\linewidth]{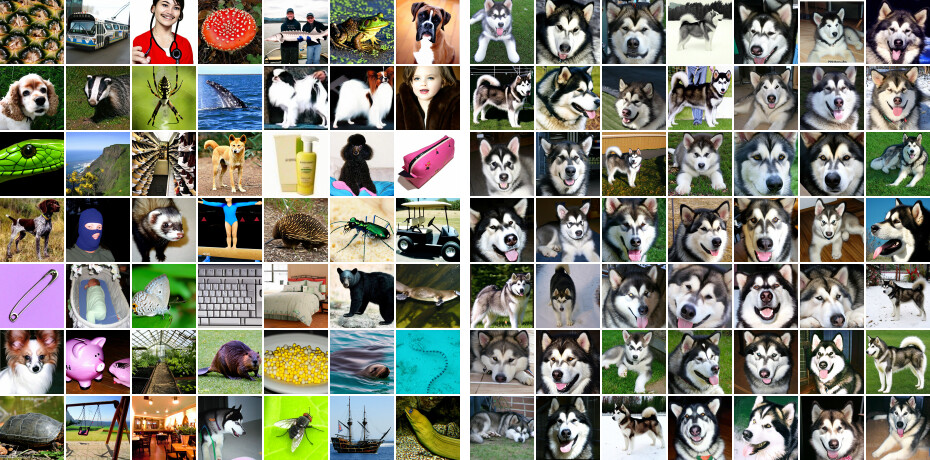}
         \caption{Classifier-free guidance with $w=3.0$: FID=24.83, IS=250.4}
     \end{subfigure}
        \caption{Classifier-free guidance on ImageNet 64x64. Left: random classes. Right: single class (malamute). The same random seed was used for sampling in each subfigure.}
        \label{fig:i64_samples}
\end{figure}

\begin{figure}[H]\vspace{-1em}
     \centering
     \begin{subfigure}[b]{\textwidth}
         \centering
         \includegraphics[width=\linewidth]{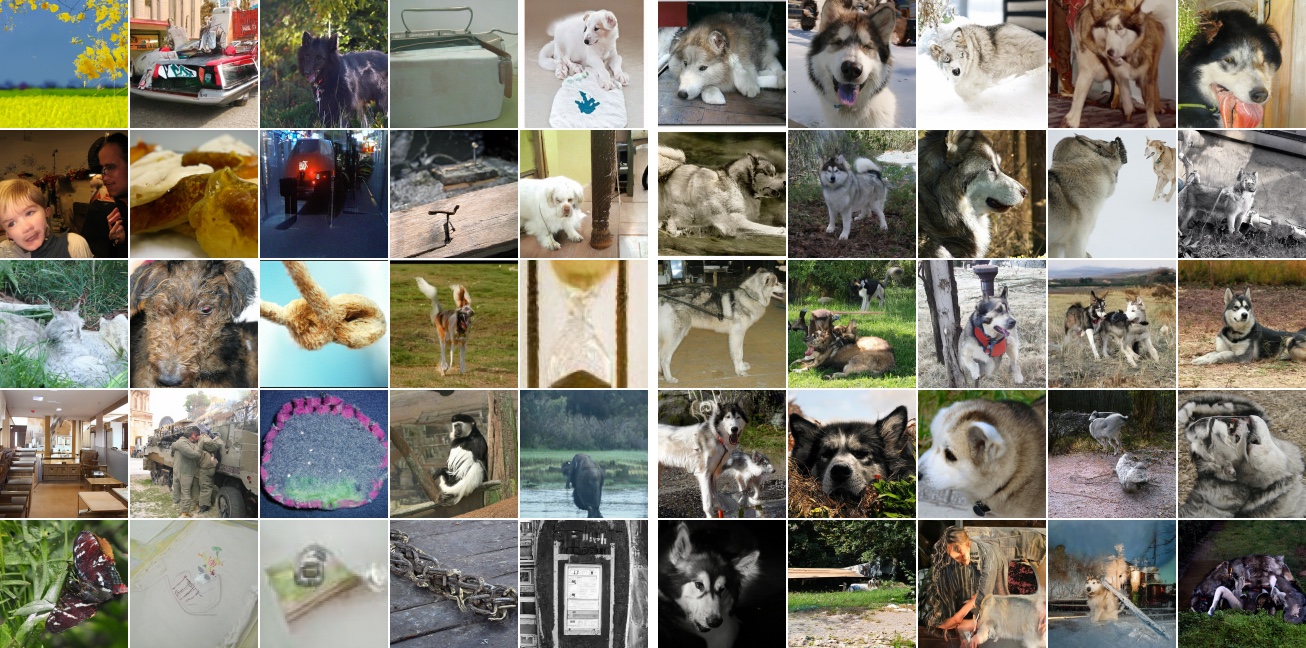}
         \caption{Non-guided conditional sampling: FID=7.27, IS=82.45}
     \end{subfigure}
     \begin{subfigure}[b]{\textwidth}
         \centering
         \includegraphics[width=\linewidth]{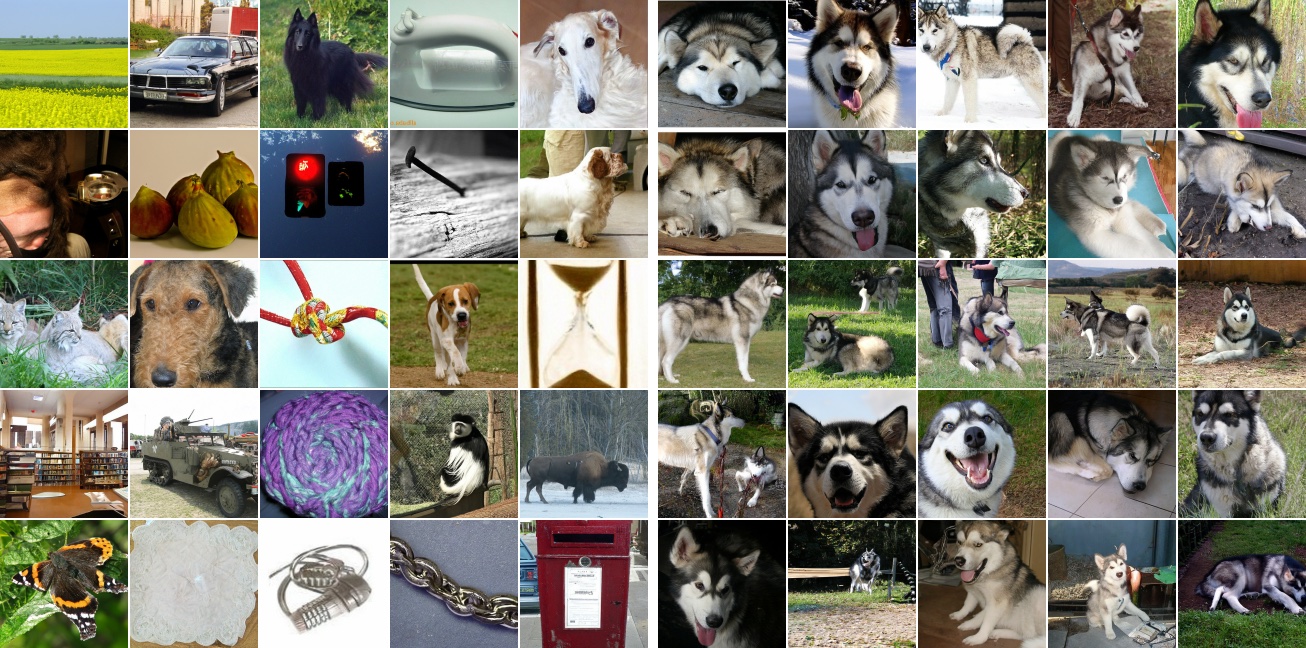}
         \caption{Classifier-free guidance with $w=1.0$: FID=7.86, IS=297.98}
     \end{subfigure}
     \begin{subfigure}[b]{\textwidth}
         \centering
         \includegraphics[width=\linewidth]{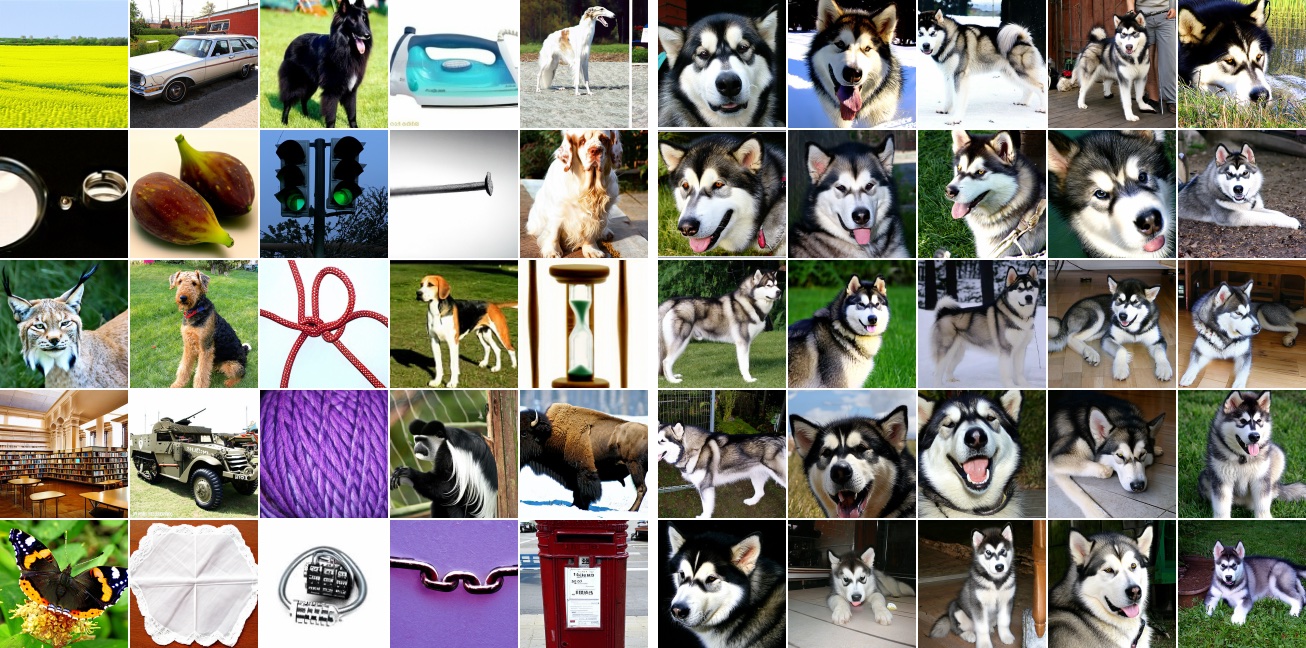}
         \caption{Classifier-free guidance with $w=4.0$: FID=21.53, IS=421.03}
     \end{subfigure}
        \caption{Classifier-free guidance on ImageNet 128x128. Left: random classes. Right: single class (malamute). The same random seed was used for sampling in each subfigure.}
        \label{fig:i128_samples}
\end{figure}

\begin{figure}[t] \centering
\includegraphics[width=\linewidth]{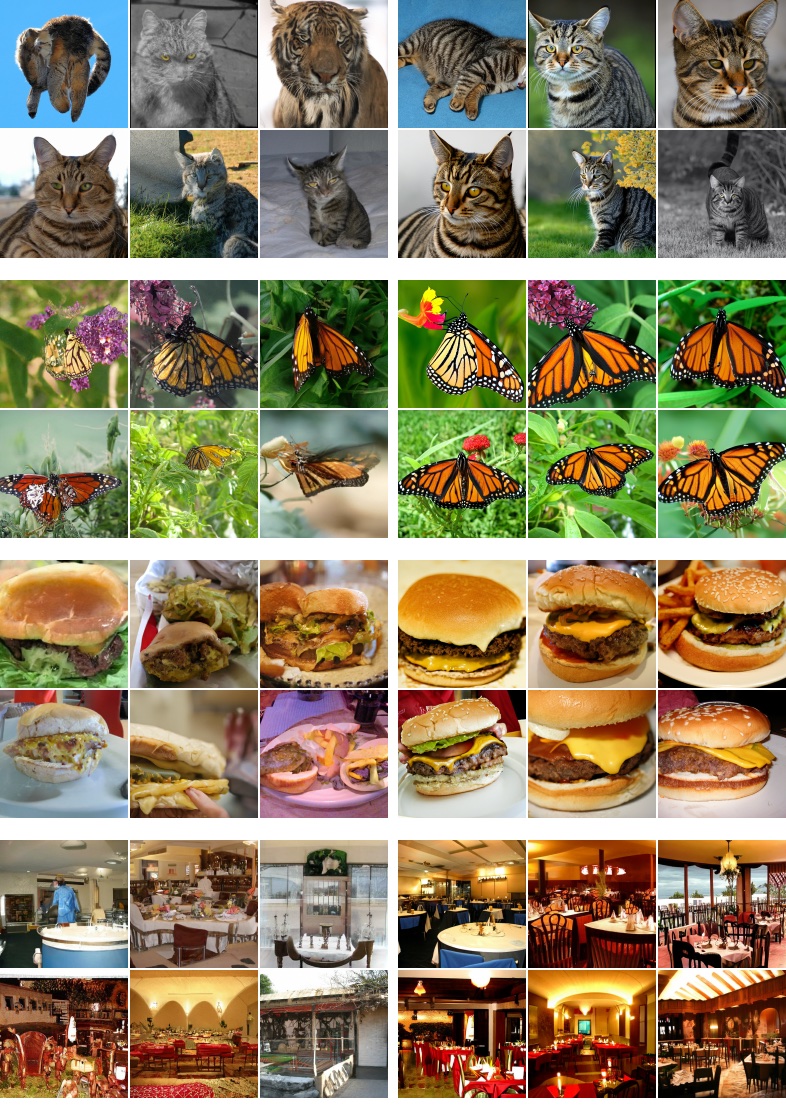}
\caption{More examples of classifier-free guidance on 128x128 ImageNet. Left: non-guided samples, right: classifier-free guided samples with $w=3.0$.}
\label{fig:i128_guidance_more}
\end{figure}

\end{document}